\documentclass[runningheads]{llncs}

\usepackage{graphicx}
\usepackage{multirow}
\usepackage{multicol}
\usepackage{xcolor}
\usepackage{url}
\usepackage{marvosym}
\begin{document}
\title{Segmentation with Multiple Acceptable Annotations: A Case Study of Myocardial Segmentation in Contrast Echocardiography}
\titlerunning{Segmentation with Multiple Acceptable Annotations}
%
\author{Dewen Zeng \inst{1}$^{(\textrm{\Letter})}$ \and
Mingqi Li \inst{2} \and
Yukun Ding \inst{1} \and
Xiaowei Xu \inst{2} \and
Qiu Xie \inst{2} \and
Ruixue Xu \inst{2} \and
Hongwen Fei \inst{2} \and
Meiping Huang \inst{2} \and
Jian Zhuang \inst{2} \and
Yiyu Shi \inst{1}
}
\authorrunning{D. Zeng, et al.}
%
\institute{
University of Notre Dame, Notre Dame, USA \\
\email{dzeng2@nd.edu}
\and
Guangdong Provincial People's Hospital, Guangzhou, China
}
\maketitle              
\begin{abstract}
Most existing deep learning-based frameworks for image segmentation assume that a unique ground truth is known and can be used for performance evaluation. This is true for many applications, but not all. Myocardial segmentation of Myocardial Contrast Echocardiography (MCE), a critical task in automatic myocardial perfusion analysis, is an example. 
Due to the low resolution and serious artifacts in MCE data, annotations from different cardiologists can vary significantly, and it is hard to tell which one is the best.
In this case, how can we find a good way to evaluate segmentation performance and how do we train the neural network?
In this paper, we address the first problem by proposing a new extended Dice to effectively evaluate the segmentation performance when multiple accepted ground truth is available.
Then based on our proposed metric, we solve the second problem by further incorporating the new metric into a loss function that enables neural networks to flexibly learn general features of myocardium.
Experiment results on our clinical MCE data set demonstrate that the neural network trained with the proposed loss function outperforms those existing ones that try to obtain a unique ground truth from multiple annotations, both quantitatively and qualitatively. 
Finally, our grading study shows that using extended Dice as an evaluation metric can better identify segmentation results that need manual correction compared with using Dice.
\end{abstract}
\section{Introduction}
Deep Neural Networks (DNNs) have been widely used in supervised image segmentation tasks, which rely on manual annotations to provide ground truth in training and evaluation 
\cite{litjens2017survey,xu2019whole,ronneberger2015u,liu2019machine}.
However, in many cases there exist large variations among different annotators due to various reasons including human factors and image qualities. 
For variations caused by human factors such as differences in annotators' training, expertise and consistency over time, \cite{sudre2019let} and  \cite{tanno2019learning} present methods to train DNNs 
to learn the behaviour of individual annotators as well as their consensus. As such, the resulting performance is much better than that can be achieved by learning from one annotator alone. 
An important assumption in these methods is that a unique ground truth is known in the evaluation process 
and can be obtained by the majority vote of the experienced annotators, which is true for human-factor induced variations.

\begin{figure}[t]
\centering
\includegraphics[width=0.9\linewidth]{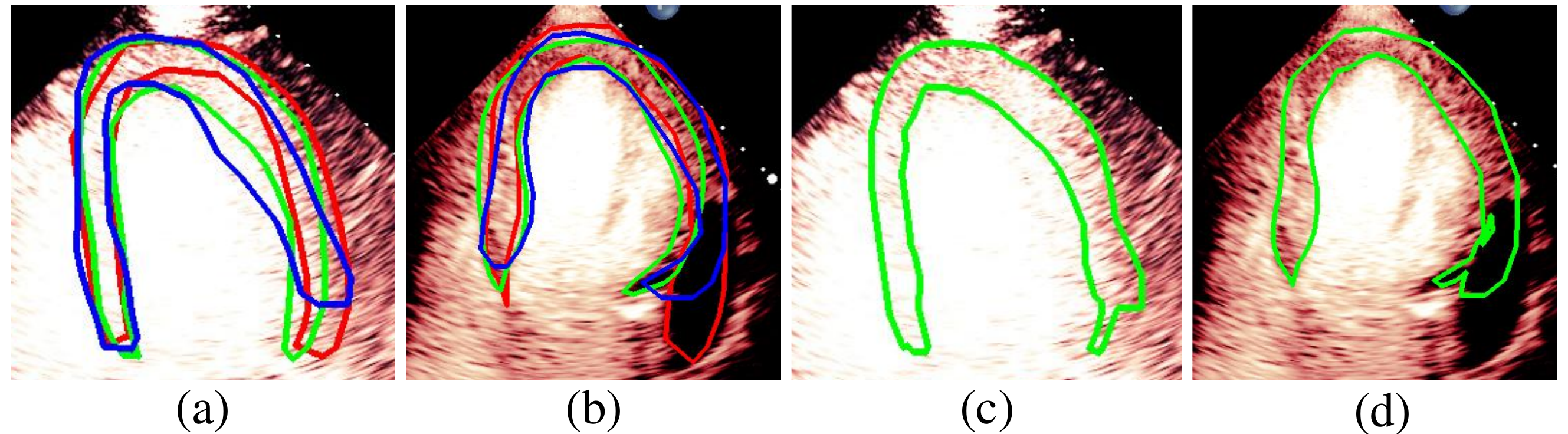}
\caption{Visualization of annotations from three experienced cardiologists marked with red, blue and green in (a) (b), and the corresponding pixel-wise majority vote (c) (d).}
\label{fig1}
\end{figure}

For variations caused by low quality images such as those of low resolution or significant artifacts, 
however, the unique ground truth may not be available.
Take the myocardial segmentation task of Myocardial Contrast Echocardiography (MCE) as an example.
An inter-observer experiment was conducted among five experienced cardiologists, and 
Fig. \ref{fig1}(a)(b) visualize the annotations of two images from three of these cardiologists.  
It can be seen that the labels by different cardiologists vary significantly, especially in locations where the intensity information of myocardium is very similar to the background. 
Table~\ref{table_1} shows the average Dice of the annotation of each cardiologist, using 
one of the others' as the ground truth, over 180 images. 
We can observe that none of the Dice is above 0.9, some even under 0.8, confirming significant variations 
among the annotations. In this case, as the variations are caused by the image quality, even these cardiologists cannot tell which annotation is better than others, and a majority vote for ground truth would not make sense sometimes as can be seen in Fig. \ref{fig1}(c)(d). 
For this reason, we cannot obtain a unique ground truth in the evaluation process and the traditional metrics such as Dice and IoU cannot be used.


\begin{table*}[t]
\caption{Average Dice of the annotations of each cardiologist using one of the others' as ground truth (calculated from 180 annotated images).}\smallskip
\centering
\smallskip
\begin{tabular}{|c|c|c|c|c|c|}
\hline
      & cardiologist 1   & cardiologist 2 & cardiologist 3 & cardiologist 4 & cardiologist 5    \\ \hline
cardiologist 1 & 1      & - & -  & - & -     \\ \hline
cardiologist 2 & 0.898 & 1 & - & - & -\\ \hline
cardiologist 3 & 0.844 & 0.849 & 1 & -  & - \\ \hline
cardiologist 4 & 0.783 & 0.790 & 0.800  & 1 & - \\ \hline
cardiologist 5 & 0.803 & 0.807 & 0.814  & 0.787 & 1  \\  \hline
\end{tabular}
\label{table_1}
\vspace{-0.1in}
\end{table*}

This leads to our key motivation, in this paper, we propose a new extended Dice metric to effectively evaluate the quality of segmentation performance when multiple accepted ground truths are available in the evaluation process.
We further incorporate the new metric into a loss function to train our segmentation network, which can help the networks better learn the general features of myocardium and ignore variations caused by individual annotators. 
To evaluate our proposed method, we collect an MCE data set of decent size with annotations from multiple experienced cardiologists. 
Experimental results on the data set show that compared with existing methods that try to create a unique ground truth in evaluation through multiple annotations, our method can achieve a higher extended Dice. Furthermore, even if we assume that the ground truth 
is one of the cardiologists' annotation or the majority voting of all cardiologists' annotations, our proposed method always 
outperforms the existing methods consistently in terms of conventional metrics such as 
Dice, Intersection over Union (IoU) and Hausdorff distance, showing stronger robustness. 
In addition, in terms of clinical value, our method also performs the best on the extraction of frame-intensity curve, commonly used in myocardial perfusion analysis \cite{porter2018clinical,dewey2020clinical}, as well as in a visual grading study of the segmentation results. 
Finally, our result shows that the proposed extended Dice can better identify segmentation results that need manual correction compared with using Dice.
In view of the lack of MCE data set available in the public domain, we will make ours available \cite{dataset}.

\section{Related Work}
Currently the analysis of MCE data heavily relies on human visual system. 
The assessment of coronary artery disease (CAD) through MCE data is based on the observation and knowledge of cardiologists, which is time consuming and hardly replicable.
Therefore, automatic myocardial segmentation of MCE can help reduce the workload of cardiologists and improve productivity.
Compared to traditional B-mode echocardiopraphy, MCE data have a few unique challenges: a) The signal-to-noise ratio is low and the contrast changes a lot over time because of the movement of microbubbles; b) The shape and pose of myocardium vary with heart motion, body physical difference, and scan setting \cite{tang2011quantitative}. Different chamber views have different myocardial structure feature; and c) Misleading structures such as papillary muscle have the same intensity and grayscale information as myocardium, which makes it harder to find the myocardium border accurately.


There exist some works that focus on training neural networks with noisy labels assuming independence between samples and noise \cite{sudre2019let,tanno2019learning,wang2012multi,warfield2004simultaneous}.
\cite{warfield2004simultaneous} proposed an algorithm to estimate the underlying true segmentation from a collection of segmentations generated by human raters or automatic algorithms. 
However, information about the original image is completely neglected.
\cite{tanno2019learning} proposed a method to simultaneously learn the individual annotator model and the underlying true label distribution through the confusion matrices of annotators.
\cite{sudre2019let} demonstrated that jointly modeling both individual and consensus estimates can lead to significant improvements in performance.
Although these methods deal with classification problems where multiple annotators exist, the variations were caused by human factors and a unique ground truth can still be obtained in the evaluation process by the majority vote of multiple experienced annotators. 
Such an approach, however, may not work well in myocardial segmentation of MCE data as the variations are caused by the image quality. 
It is hard to tell the best one among multiple annotations from experienced cardiologists or to perform pixel-wise majority vote which may lead to irregular boundaries. 
In other words, even in the evaluation process, it is impossible to obtain the ground truth. 

\section{Method}

In this section, we introduce a new metric for image segmentation tasks where multiple acceptable annotations exist in the evaluation process. 
According to our observation on the MCE data, the annotation from each cardiologist is acceptable clinically, i.e., it can be directly used in myocardial perfusion analysis. 
Such an observation is also true in many other medical applications such as CT measurement \cite{mcerlean2013intra} and MRI interpretation \cite{beresford2006inter}.
Note that this is fundamentally different from human-factor caused variations, which may lead to noisy or error-prone annotations. They can be addressed using methods discussed in \cite{tanno2019learning,sudre2019let} and thus are not included in our discussion.

\begin{figure*}[t]
  \centering
  \includegraphics[width=1.0\columnwidth]{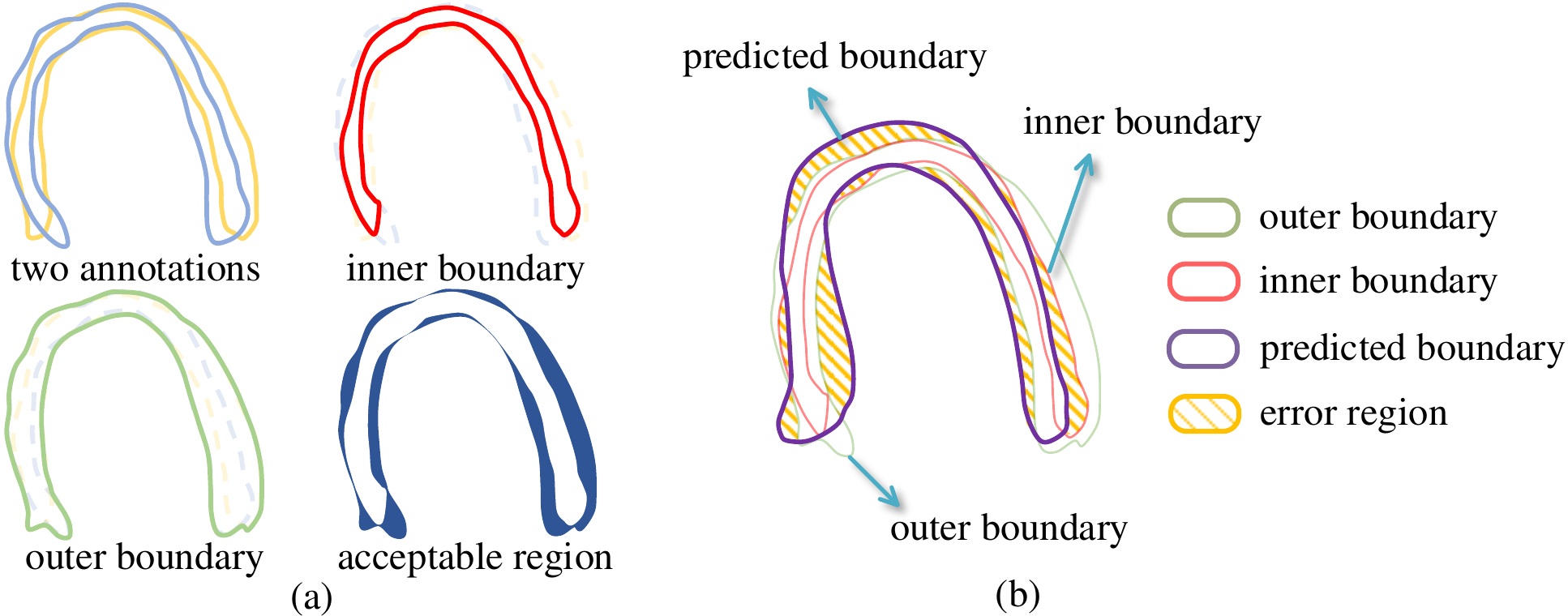}
  \caption{(a) Visual illustration of two different boundary annotations, inner boundary, outer boundary and acceptable region, respectively. (b) Illustration of the computation of extended Dice, a toy example.}
  \label{fig2}
  \vspace{-0.2in}
\end{figure*}

For simplicity of discussion, we use myocardial boundary annotations from two cardiologists as an example. The same concept can be readily extended to more cardiologists. 
We first obtain an inner boundary and an outer boundary based on these two annotations, as shown in Fig. \ref{fig2}(a).
The inner boundary is defined by that of the intersection of the two regions enclosed by the annotated boundaries. In other words, every pixel inside the inner boundary is labeled as myocardium by both of the cardiologists. 
The outer boundary is that of the union of the two regions, i.e., every pixel inside the outer boundary is labeled as myocardium by at least one cardiologist.
As both annotations are acceptable based on our assumption, we can obtain an acceptable region, which is the region between the inner boundary and outer boundary. 
The pixels inside the acceptable region can be classified as either myocardium or background; any boundary that completely falls inside the acceptable region shall be considered acceptable.
Intuitively, in light of the existence of multiple acceptable annotations, the introduction of acceptable region can allow the prediction boundary to have some flexibility in regions where significant large inter-observer variability exists. 
Such flexibility will potentially lead to better overall segmentation quality.

Based on the observation above, we can extend the traditional Dice metric which is based on a single ground truth. 
Denote the region enclosed by the inner boundary and the outer boundary as $I$ and $O$, respectively. The region inside the predicted boundary
is $P$. Then our metric can be calculated as

\begin{equation}
Metric(P,I,O) = 1-\frac{(P-P\cap O)+(I-P\cap I)}{P+I}.
\end{equation}

A simple illustration is shown in Fig. \ref {fig2}(b). 
We can easily see that the metric is penalizing the total segmented region outside the outer boundary $P-P\cap O$ and inside the inner boundary $I-P\cap I$.
The denominator $P+I$ helps scale the final result to range $[0,1]$. 
We can see that if the predicted boundary is completely inside the acceptable region, the metric is 1. Otherwise, if it has no overlap with the acceptable region, the metric is 0.

{\bf Relationship to Dice:} Notice that our metric can be simplified into
\begin{equation}
Metric(P,I,O) = \frac{(P\cap O)+(P\cap I)}{P+I}.
\end{equation}
When multiple annotators give the same annotation results, which means the ground truth annotation is known and the inner boundary and the outer boundary completely overlap, the proposed metric will become the conventional Dice. Therefore, it can be viewed as extended Dice. 




Once we have the proposed metric, we can further use it as a loss function to train a neural network to help the model focus more on the general features of myocardium and ignore variations from the individual cardiologist for better performance. 
We denote the pixel-wise annotation of the regions enclosed by the inner boundary and the outer boundary as $I_{n}$ and $O_{n}$, and the prediction for the myocardium region in the training process as $P_{n}$ $\in{[0,1]}$, respectively. 
$n$ is the index of pixel space $N$. Then our extended Dice loss function is then defined as follows:
\begin{equation}
\mathcal{L} = \frac{\sum_{n=1}^{N}(P_{n} - (P_{n}\times{O_{n}}))+\sum_{n=1}^{N}(I_{n} - (I_{n}\times{P_{n}}))}{\sum_{n=1}^{N}(P_{n}+I_{n})}.
\end{equation}



The idea of the introduced loss function is that, for pixels inside the acceptable region, any prediction result is reasonable and should not  contribute to the loss function. For pixels outside the acceptable region, any misclassification should be penalized. 
\section{Experiment}

\subsection{Dataset}


Our dataset consists of 100 subjects in total, 40 of which were diagnosed with coronary artery diseases (CAD) and the rest were not. MCE sequence data was collected from these subjects by an ultrasonography system (Philips 7C or iE ELITE, Philips Medical Systems, Best, Netherlands) equipped with a broadband transducer, and SonoVue (Bracco Research SA, Geneva, Switzerland) as the contrast agent. For each of the subjects, an MCE sequence in the apical of 4-chamber view was acquired, and we randomly selected 10 images from the MCE sequence, which result in an MCE dataset with 1000 images in total.
We split our data into training and test set with a ratio of 7:3, i.e., 700 images from 70 subjects were used for training and 300 images from 30 subjects were used for validation. 
The manual annotations of myocardium in MCE images were performed by five experienced cardiologists, and the time for labeling each image is around 1-2 minutes per cardiologist. 
Our dataset is available online at \cite{dataset}.


\subsection{Training DNN with Extended Dice Loss}

\subsubsection{Experiment Setup}

We base our experiments on two segmentation networks, U-Net and DeepLab, using Pytorch based implementations in \cite{isensee2018nnu} and \cite{chen2018encoder}, respectively.
The initial MCE images were cropped into $512\times{512}$. 
For data augmentation during training, we randomly scale all the images by $[0.8,1.2]$ and rotate them by $[-30^{\circ},30^{\circ}]$. We also shifted the image brightness, contrast and saturation by $[-0.1,0.1]$. During test, we do not employ any augmentations.
Batch size is set to 5 and the training epoch is 20. 
The learning rate is 0.0002 for the first half of epochs, and then 0.00002 for the rest.

We train U-Net and DeepLab using the proposed loss function and compare it with the following methods. For training with a single annotation as the ground truth, we include the following (a) one of the cardiologists, (b) the inner boundary, (c) the outer boundary and (d) the consensus boundary of all cardiologists through pixel-wise majority vote, which are referred to as Single Cardiologist (SC), Inner Boundary (IB), Outer Boundary (OB), and Consensus, respectively. Cross-entropy loss is used in these methods. 
For training with multiple annotations, we adopt the average cross-entropy of all cardiologists referred as Average Cross Entropy (ACE) and three state-of-the-art approaches refered as Confusion Matrix (CM) \cite{sudre2019let}, Consistency \cite{tanno2019learning} and STAPLE \cite{butakoff2007simulated}. 
For STAPLE, we use the fast implementation from \cite{yaniv2018simpleitk}.
Note that these four methods, while taking multiple annotations into consideration in the training, still assume that a unique ground truth is known in the evaluation through majority voting. 
For example, in \cite{tanno2019learning} ground truth is acquired by choosing the samples where the three most experienced sonographers agreed in a given label in their cardiac view classification experiments.
However, in our problem the unique ground truth is unknown.

\begin{table*}[t]
\centering
\caption{Performance evaluation of different methods using Dice, IoU and Hausdorff distance (pixel point), respectively. Ground truth (GT) is assumed to be one of the cardiologists' annotations (cardiologist 1 to cardiologist 5) or the majority vote of five cardiologists' annotations. Results are reported in the form of mean(standard deviation) of all test images.}
\resizebox{1.0\textwidth}{!}{
\begin{tabular}{lccccccccc}
\hline
\multirow{2}{*}{Method}         & \multicolumn{3}{c}{GT: cardiologist 1} & \multicolumn{3}{c}{GT: cardiologist 2} & \multicolumn{3}{c}{GT: cardiologist 3} \\
 & Dice & IoU & HD & Dice & IoU & HD & Dice & IoU & HD \\ \hline
Single Cardiologist & 0.760(.09) & 0.623(.10) & 35.4(15) & 0.809(.13) & 0.694(.12) & 32.8(17) & 0.818(.11) & 0.707(.13) & 30.5(16) \\ 
Inner Boundary & 0.732(.10) & 0.589(.10) & 37.1(18) & 0.741(.11) & 0.601(.11) & 37.9(17) & 0.744(.11) & 0.605(.11) & 36.0(17) \\ 
Outer Boundary & 0.729(.07) & 0.581(.09) & 40.1(12) & 0.782(.10) & 0.654(.12) & 36.4(13) & 0.790(.10) & 0.665(.12) & 36.1(12) \\ 
Consensus & 0.765(.10) & 0.632(.11) & 35.3(18) & 0.824(.13) & 0.719(.14) & 31.7(19) & 0.833(.13) & 0.732(.14) & 30.0(18) \\ 
Average Cross Entropy & 0.765(.09) & 0.631(.11) & 33.2(14) & 0.819(.12) & 0.709(.13) & 29.4(15) & 0.827(.12) & 0.722(.13) & 28.0(14) \\ 
Confusion Matrix \cite{tanno2019learning} & 0.752(.10) & 0.614(.11) & 45.2(21) & 0.808(.13) & 0.695(.14) & 40.4(21) & 0.817(.13) & 0.708(.14) & 39.0(21) \\ 
Consistency \cite{sudre2019let} & 0.770(.08) & 0.635(.10) & 37.2(15) & 0.826(.11) & 0.719(.13) & 32.3(16) & 0.831(.11) & 0.726(.13) & 31.7(16) \\ 
STAPLE \cite{warfield2004simultaneous} & 0.743(.07) & 0.598(.09) & 39.6(13) & 0.810(.10) & 0.694(.12) & 33.0(17) & 0.816(.11) & 0.703(.13) & 32.7(17) \\ \hline
Our method & \textbf{0.780(.08)} & \textbf{0.649(.10)} & \textbf{31.5(12)} & \textbf{0.829(.10)} & \textbf{0.721(.12)} & \textbf{28.9(14)} & \textbf{0.836(.10)} & \textbf{0.732(.12)} & \textbf{27.7(14)} \\ 
\hline
\end{tabular}}
\newline
\vspace*{0.2 cm}
\newline
\resizebox{1.0\textwidth}{!}{
\begin{tabular}{lccccccccc}
\hline
\multirow{2}{*}{Method}         & \multicolumn{3}{c}{GT: cardiologist 4} & \multicolumn{3}{c}{GT: cardiologist 5} & \multicolumn{3}{c}{GT: majority vote} \\
 & Dice & IoU & HD & Dice & IoU & HD & Dice & IoU & HD \\ \hline
Single Cardiologist & \textbf{0.824(.10)} & \textbf{0.716(.13)} & 29.4(17) & 0.791(.08) & 0.665(.10) & 33.3(17) & 0.838(.11) & 0.735(.12) & 28.4(17) \\ 
Inner Boundary & 0.766(.11) & 0.634(.12) & 36.6(17) & 0.734(.08) & 0.590(.09) & 39.1(16) & 0.770(.11) & 0.638(.11) & 34.2(17) \\ 
Outer Boundary & 0.749(.08) & 0.608(.10) & 35.5(12) & 0.800(.05) & 0.672(.07) & 32.7(12) & 0.785(.09) & 0.656(.11) & 34.0(12) \\ 
Consensus & 0.818(.12) & 0.710(.14) & 31.5(19) & 0.806(.08) & 0.686(.10) & 32.2(18) & 0.847(.12) & 0.753(.14) & 28.0(19) \\ 
Average Cross Entropy & 0.816(.11) & 0.704(.13) & 29.4(15) & 0.806(.07) & 0.685(.09) & 30.4(14) & 0.844(.11) & 0.745(.13) & 26.4(15) \\
Confusion Matrix \cite{tanno2019learning} & 0.795(.11) & 0.675(.13) & 39.7(22) & 0.805(.07) & 0.682(.09) & 38.4(19) & 0.826(.12) & 0.719(.13) & 37.9(22) \\ 
Consistency \cite{sudre2019let} & 0.816(.10) & 0.704(.13) & 31.5(15) & 0.815(.07) & 0.696(.09) & 31.5(13) & 0.847(.10) & 0.749(.13) & 29.8(16) \\ 
STAPLE \cite{warfield2004simultaneous} & 0.779(.09) & 0.647(.11) & 33.8(15) & 0.810(.04) & 0.686(.06) & 31.3(13) & 0.814(.09) & 0.695(.11) & 31.8(16) \\ \hline
Our method & 0.822(.09) & 0.710(.12) & \textbf{28.6(13)} & \textbf{0.817(.07)} & \textbf{0.699(.09)} & \textbf{29.3(13)} & \textbf{0.855(.10)} & \textbf{0.759(.12)} & \textbf{25.4(14)} 
\\ 
\hline
\end{tabular}}
\newline
\label{table_2}
\vspace{-0.2in}
\end{table*}

\subsubsection{Quantitative Analysis}

As discussed earlier, because a unique ground truth cannot be obtained through either majority vote or best of annotations, in order to evaluate the performance of our proposed method, we conduct comparisons from the following three aspects. 
(a) we treat one of the cardiologists' annotation as well as the majority vote of the annotations as the ground truth of the test images (which is again not necessarily the ``real'' ground truth), and evaluate the conventional metrics (e.g., Dice) of different methods.
(b) we compare the extended Dice across all the test images of all the methods in Table \ref{table_3}. 
(c) we use an important method named frame-intensity curve, which is commonly used for myocardial perfusion analysis in MCE \cite{porter2018clinical}, along with visual grading study by an experienced cardiologist to further show the clinical efficacy of our method.

\textbf{Evaluation using conventional metrics:}
Note that in our experiment we cannot obtain the unique ground truth.
So in order to use traditional evaluation metric to assess the model performance, we assume that one of the cardiologists' annotations or the majority vote of the annotations is the ground truth in the test set and compute the conventional Dice, IoU and Hausdorff distance of the segmentation result of different methods.
U-net is used as the network architecture.
The results can be seen in Table \ref{table_2}. Note that only the best result among the five cardiologists is reported for the ``single cardiologist'' method.
We can observe that among all the methods, our method performs best consistently (i.e., with highest Dice/IoU, lowest Hausdorff distance and the smallest standard deviation) when any of the cardiologists' annotation is used as the ground truth.
The conclusion also holds if we use the majority voting of all cardiologists' annotations as the ground truth. As such, our method shows stronger robustness over other methods. 
Notice that when using cardiologist 4 as the ground truth, the Single cardiologist method performs best in terms of Dice coefficient and IoU. This is because in this particular method the model is trained with cardiologist 4's annotation as the ground truth.
However, the method's performance drops significantly when other cardiologists' annotations are used as the ground truth.


\begin{table}[t]
\centering
\caption{Performance comparison in terms of extended Dice using different methods. Results are reported as mean(standard deviation) across all test images. SC, IB, OB, ACE and CM refer to Single Cardiologist, Inner Boundary, Outer Boundary, Average Cross Entropy and Confusion Matrix, respectively.}
\resizebox{1.0\textwidth}{!}{
\begin{tabular}{llllllllll}
\hline
Method & SC & IB & OB & Consensus & ACE & CM & Consistency & STAPLE & Ours \\ \hline
U-Net & 0.929(.06) & 0.947(.05) & 0.848(.07) & 0.940(.06) & 0.919(.06) & 0.951(.06) & 0.947(.06) & 0.912(.06) & \textbf{0.958(.05)} \\
DeepLab & 0.942(.07) & 0.906(.08) & 0.891(.07) & 0.946(.08) & 0.945(.07) & 0.924(.08) & 0.944(.07) & 0.921(.07) & \textbf{0.954(.06)} \\
\hline
\end{tabular}
}
\label{table_3}
\end{table}

\textbf{Evaluation using extended Dice:} We then evaluate these methods using the proposed extended Dice metric, which does not require the unique ground truth to be known. 
Specifically, the mean/standard deviation of the extended Dice of all methods are calculated.
U-net and Deeplab architecture are used, we use the same experimental setup as discussed in Section 4.2.
From Table \ref{table_3} we can see that our method always achieves the highest extended Dice among all the methods on both U-Net and DeepLab. 
The standard deviation of our method is also lower than the others, which shows our method outperforms the others statistically.
This is because, in our training approach, the acceptable region allows neural networks some flexibility to learn more general texture and structure features of myocardium. 
Notice that Confusion Matrix and Consistency methods target classification problems originally, so their performance in the segmentation task may not be as good.


\begin{figure}[t]
  \centering
  \includegraphics[width=0.9\columnwidth]{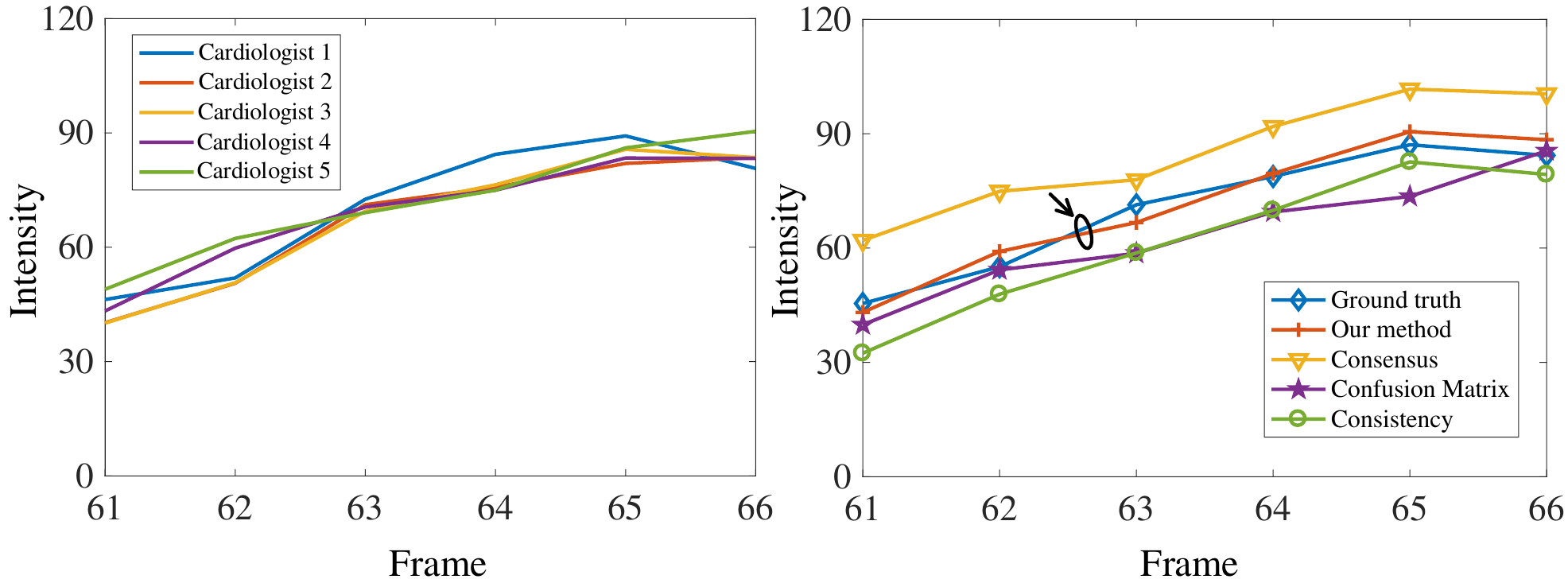}
  \caption{Quantitative analysis of different methods. Left: frame-intensity curves from the annotations of five different cardiologists, which are very close to each other. Right: frame-intensity curves of different methods and the ground truth obtained by averaging the curves on the left. Ours is the closest to the ground truth.}
  \label{figure_3}
  \vspace{-0.2in}
\end{figure}

\textbf{Evaluation using frame-intensity curve:} In order to show that the proposed extended Dice and loss function are indeed superior from the clinical application perspective, we also use the frame-intensity curve of segmented myocardium, which is commonly used in myocardial perfusion analysis \cite{porter2018clinical,dewey2020clinical}, to evaluate the segmentation performance.
Frame-intensity curve, also known as the time-intensity curve, can be used to reflect the relative microvascular blood volume after microbubbles infusion, thus help estimate myocardial ischemia.
Fig. \ref{figure_3} shows the frame-intensity curves of six frames of a subject (Frame 61-66) using the annotations from each of the five cardiologists (left), and from the top four methods in Table \ref{table_3}, namely our method, Consensus, Confusion Matrix and Consistency (right). U-Net is used as the segmentation framework.  
Although the annotations from the five cardiologists are very different as shown in Table \ref{table_1}, from the left figure we can see that the resulting frame-intensity curve from each of them are very similar to each other. 
Based on this observation, we obtain a ground truth frame-intensity curve through averaging.
It can be seen in the right figure that among all the four methods compared, the curve generated by our method is the closest to the ground truth curve. 
This convincingly shows that our method can better help myocardial perfusion analysis. 

\subsubsection{Qualitative Analysis}

\begin{table}[t]
\centering
\caption{Visual grading study of different methods by an independent and experienced cardiologist without knowing which method is applied on each image.}
\resizebox{0.85\columnwidth}{!}{
\begin{tabular}{lcccc}
\hline
Grading Level & Consensus & Confusion Matrix & Consistency & Our method \\ \hline 
Level 4 (Highest)       & 58 &   71                 &      69          &        \textbf{72}          \\ 
Level 3           & 56 &   39       &      42          &         50         \\ 
Level 2        & 11 &   17    &      22          &          20         \\ 
Level 1 (Lowest)   & 25 & 23  &      17         &         \textbf{8}         \\ \hline
\end{tabular}}
\label{table_4}
\end{table}

\begin{figure*}[t]
\centering
\resizebox{0.9\textwidth}{!}{
\begin{tabular}{cccccc}
Original image & Consensus & Confusion Matrix & Consistency & Our method \\
    \includegraphics[width=0.2\textwidth]{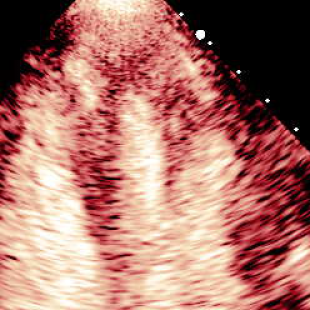} &
    \includegraphics[width=0.2\textwidth]{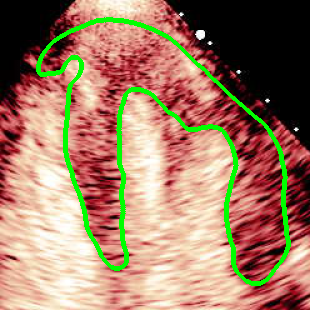} &  \includegraphics[width=0.2\textwidth]{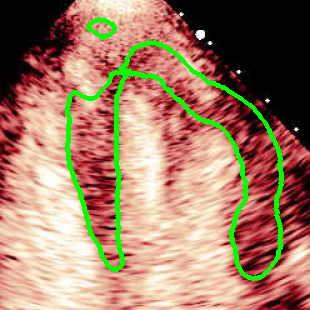} & 
    \includegraphics[width=0.2\textwidth]{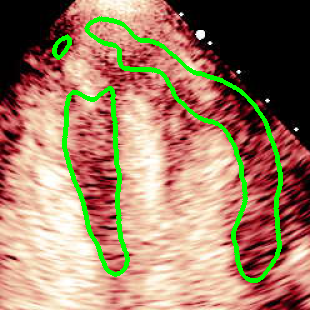} & 
    \includegraphics[width=0.2\textwidth]{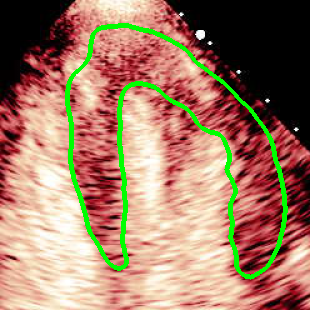}  \\
    \includegraphics[width=0.2\textwidth]{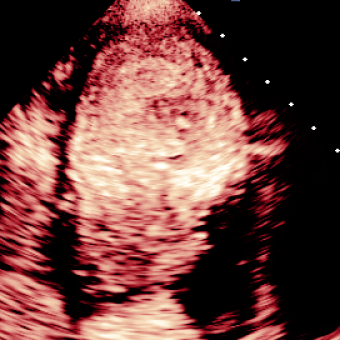} &
    \includegraphics[width=0.2\textwidth]{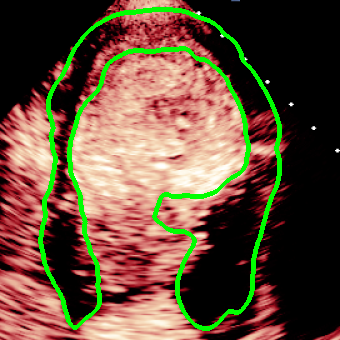} &  \includegraphics[width=0.2\textwidth]{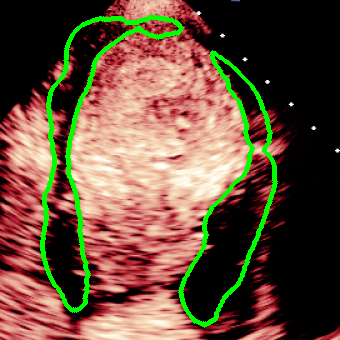} & 
    \includegraphics[width=0.2\textwidth]{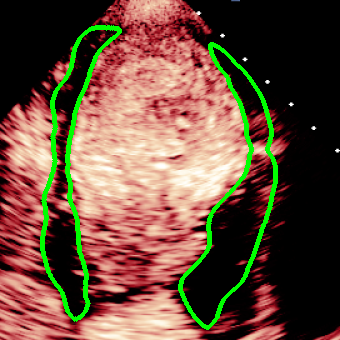} & 
    \includegraphics[width=0.2\textwidth]{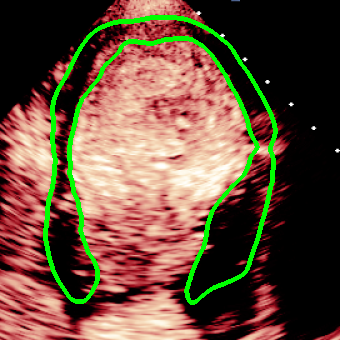}  \\
\end{tabular}}
\caption{Visualization of myocardial segmentation results of two examples using U-Net as the backbone with different methods.}
\label{figure_5}
\vspace{-0.2in}
\end{figure*}

A visual grading study is conducted to further demonstrate the efficacy of our proposed metric and the loss function.
In this experiment, an independent and experienced cardiologist is asked to grade the myocardial segmentation result in a blind setting (i.e., without knowing which method was used).
We randomly select 5 frames for each subject in the test set for grading, resulting in 30$\times$5 frames in total. 
Grading is based on the segmentation accuracy which can be classified into 4 levels: Level 4, excellent - no manual correction needed; Level 3, slight yet observable shape deviation; Level 2, obviously irregular shapes; Level 1, severe disconnection/mistakes.
Similar to the frame-intensity curve study, we compare our method with the three best methods in Table \ref{table_3}. 
Table \ref{table_4} demonstrates the grading results for each method.
It can be seen that compared with other methods, our proposed method has the most number of frames in the highest quality level (level 4), and fewest in the lowest level (level 1), which means that  out method is less likely to made a segmentation which has severe disconnections or mistakes.


Visualization of myocardial segmentation from these methods is also shown in Fig. \ref{figure_5}. 
We can see that for the Consensus, Confusion Matrix and Consistency methods, some unregulated shapes or discontinuity exist because the model cannot discriminate the actual myocardial from the artifacts which have the same intensity and texture information.
However, the results of our method can alleviate the problem and accurately segment myocardium in MCE images.

\begin{figure}[t]
  \centering
  \includegraphics[width=0.55\columnwidth]{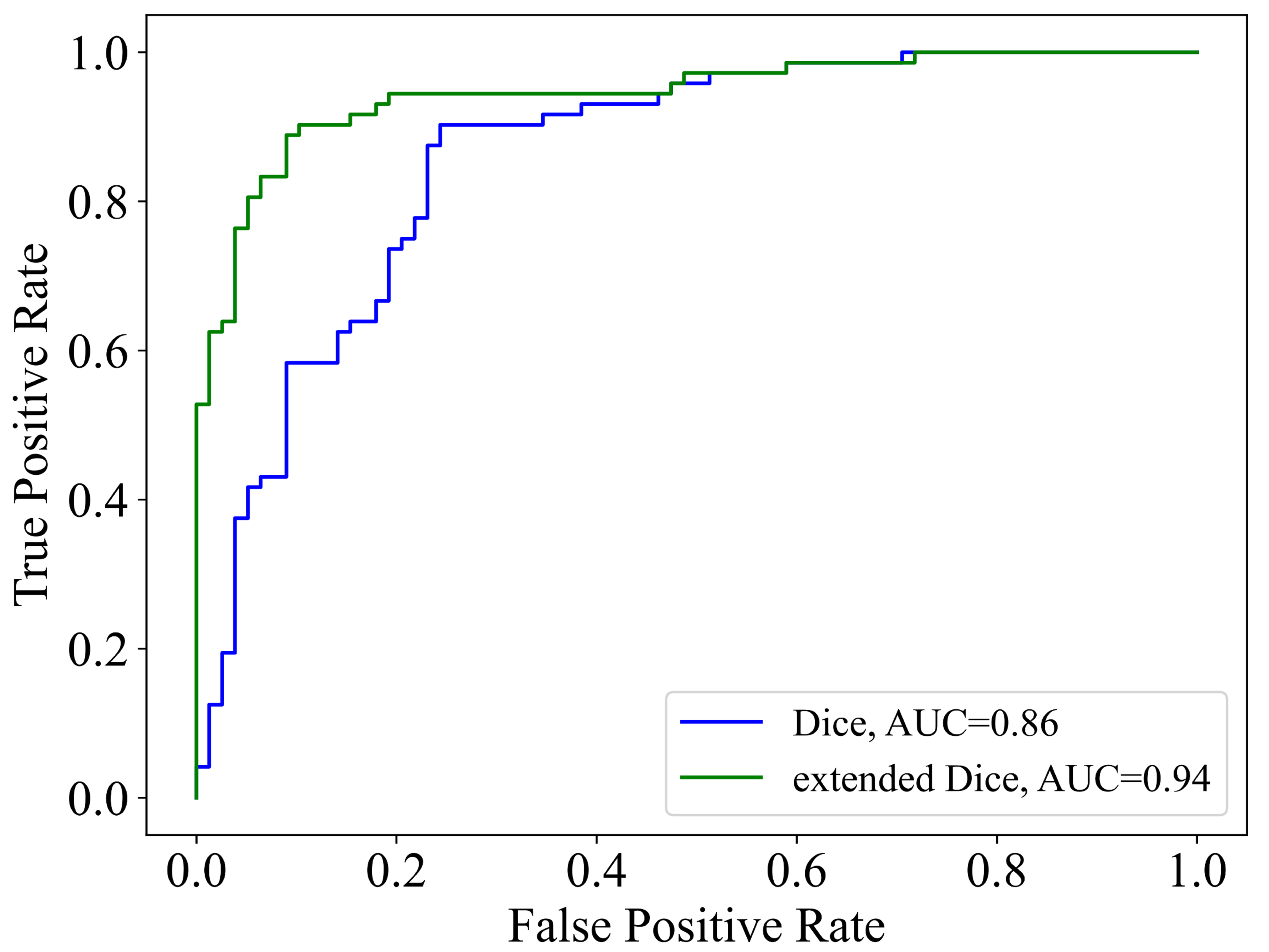}
  \caption{ROC curve of using Dice and extended Dice as the segmentation evaluation metric in classifying segmentations based on whether manual correction is needed.}
  \label{figure_6}
  \vspace{-0.2in}
\end{figure}

\subsection{Extended Dice as a Superior Evaluation Metric}

In order to demonstrate the advantage of extended Dice over Dice as an evaluation metric, we use these two metrics to evaluate the prediction generated by our training method in the grading study in Section 4.2.
The objective is to identify the segmentation results that need manual correction, which can be viewed as a binary classification problem.
The segmentations in the grading Level 1, Level 2 and Level 3 are defined as the class 0 that need manual correction.
The segmentations in Level 4 are defined as class 1 which do not need manual correction.
Fig. \ref{figure_6} shows the Receiver Operating Characteristic (ROC) curve of using Dice and extended Dice as the metric for the classification problem.
It can be seen that using extended Dice, we can improve the classification AUC from 0.86 to 0.94, suggesting that compared to Dice, extended Dice is better in distinguishing segmentations that need manual correction from those that do not.

\begin{figure}[t]
  \centering
  \includegraphics[width=0.95\columnwidth]{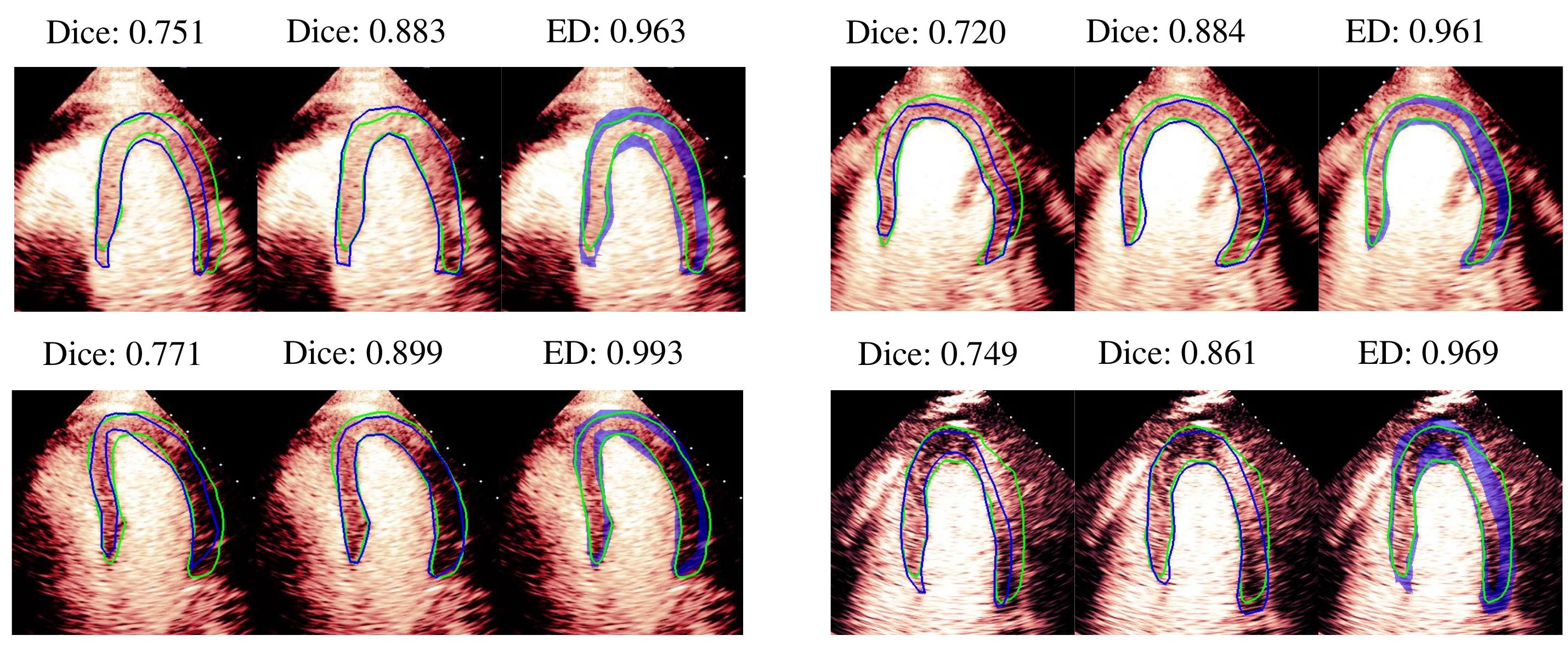}
  \caption{Evaluation comparison of four images using Dice and extended Dice (denoted as ED). Images from left to right in one group represent the Dice evaluated by cardiologist 2, Dice evaluated by cardiologist 5, and extended Dice evaluated by all cardiologists, respectively. Green boundary is the prediction, blue boundary is the ground truth, and blue area is the acceptable region.}
  \label{figure_7}
  \vspace{-0.2in}
\end{figure}

In Fig. \ref{figure_7}, we show the evaluation comparison of four images using Dice and extended Dice.
Images in each group represent the Dice evaluated by cardiologist 2, Dice evaluated by cardiologist 5 and extended Dice.
It can be seen that there exist large variations in Dice when using different cardiologists' annotations as the ground truth, 
which may lead to misjudgment during evaluation.
For example, we may think the segmentation quality is low (Dice$<$0.77) based on the annotation from one of the cardiologists, but the quality is actually pretty good based on the annotation from another one (Dice$>$0.86).
However, using extended Dice will alleviate the problem because all these images are considered good (extended Dice$>$0.96).




\section{Conclusion}
In this paper, we proposed an extended Dice metric for evaluating the performance of image segmentation where multiple annotations exist and unique ground truth is unknown due to low image qualities. 
Based on the metric, we further use it as a loss function for training DNN to help the network better learn the general features of myocardium and ignore variations caused by individual annotators.
Experiments on MCE data set demonstrate that the proposed loss function can improve the segmentation performance evaluated by conventional metrics, by the proposed metric, by the extraction of frame-intensity curve used in myocardial perfusion analysis, and by a visual grading study of the segmentation results in a blind setting. 
Comparing Dice and extended Dice as segmentation evaluation metrics, extended Dice performs better in identifying images that need manual correction. 
While we only demonstrate the efficacy of the proposed method on myocardial segmentation, it is likely that it can be applied to other problems where multiple acceptable annotations are available.
%
%
%
\bibliographystyle{splncs04}
\bibliography{reference}

\begin{thebibliography}{10}
\providecommand{\url}[1]{\texttt{#1}}
\providecommand{\urlprefix}{URL }
\providecommand{\doi}[1]{https://doi.org/#1}

\bibitem{dataset}
Mce dataset. \url{https://github.com/dewenzeng/MCE_dataset}

\bibitem{beresford2006inter}
Beresford, M.J., Padhani, A.R., et~al.: Inter-and intraobserver variability in
  the evaluation of dynamic breast cancer mri. Journal of Magnetic Resonance
  Imaging: An Official Journal of the International Society for Magnetic
  Resonance in Medicine  \textbf{24}(6),  1316--1325 (2006)

\bibitem{butakoff2007simulated}
Butakoff, C., Balocco, S., Ordas, S.: Simulated 3d ultrasound lv cardiac images
  for active shape model training. In: Medical Imaging 2007: Image Processing.
  vol.~6512, p. 65123U. International Society for Optics and Photonics (2007)

\bibitem{chen2018encoder}
Chen, L.C., Zhu, Y., Papandreou, G., Schroff, F., Adam, H.: Encoder-decoder
  with atrous separable convolution for semantic image segmentation. In:
  Proceedings of the European conference on computer vision (ECCV). pp.
  801--818 (2018)

\bibitem{dewey2020clinical}
Dewey, M., Siebes, M., Kachelrie{\ss}, M., Kofoed, K.F., Maurovich-Horvat, P.,
  Nikolaou, K., Bai, W., et~al.: Clinical quantitative cardiac imaging for the
  assessment of myocardial ischaemia. Nature Reviews Cardiology
  \textbf{17}(7),  427--450 (2020)

\bibitem{isensee2018nnu}
Isensee, F., Petersen, J., et~al.: nnu-net: Self-adapting framework for
  u-net-based medical image segmentation. arXiv preprint arXiv:1809.10486
  (2018)

\bibitem{litjens2017survey}
Litjens, G., Kooi, T., Bejnordi, B.E., Setio, A.A.A., Ciompi, F., Ghafoorian,
  M., Van Der~Laak, J.A., Van~Ginneken, B., S{\'a}nchez, C.I.: A survey on deep
  learning in medical image analysis. Medical image analysis  \textbf{42},
  60--88 (2017)

\bibitem{liu2019machine}
Liu, Z., Xu, X., Liu, T., Liu, Q., Wang, Y., Shi, Y., Wen, W., Huang, M., Yuan,
  H., Zhuang, J.: Machine vision guided 3d medical image compression for
  efficient transmission and accurate segmentation in the clouds. In:
  Proceedings of the IEEE Conference on Computer Vision and Pattern
  Recognition. pp. 12687--12696 (2019)

\bibitem{mcerlean2013intra}
McErlean, A., Panicek, D.M., Zabor, E.C., Moskowitz, C.S., Bitar, R., Motzer,
  R.J., Hricak, H., Ginsberg, M.S.: Intra-and interobserver variability in ct
  measurements in oncology. Radiology  \textbf{269}(2),  451--459 (2013)

\bibitem{porter2018clinical}
Porter, T.R., Mulvagh, S.L., Abdelmoneim, S.S., Becher, H., et~al.: Clinical
  applications of ultrasonic enhancing agents in echocardiography: 2018
  american society of echocardiography guidelines update. Journal of the
  American Society of Echocardiography  \textbf{31}(3),  241--274 (2018)

\bibitem{ronneberger2015u}
Ronneberger, O., Fischer, P., Brox, T.: U-net: Convolutional networks for
  biomedical image segmentation. In: International Conference on Medical image
  computing and computer-assisted intervention. pp. 234--241. Springer (2015)

\bibitem{sudre2019let}
Sudre, C.H., Anson, B.G., et~al.: Let’s agree to disagree: Learning highly
  debatable multirater labelling. In: International Conference on Medical Image
  Computing and Computer-Assisted Intervention. pp. 665--673. Springer (2019)

\bibitem{tang2011quantitative}
Tang, M.X., Mulvana, H., Gauthier, T., Lim, A., Cosgrove, D., Eckersley, R.,
  Stride, E.: Quantitative contrast-enhanced ultrasound imaging: a review of
  sources of variability. Interface focus  \textbf{1}(4),  520--539 (2011)

\bibitem{tanno2019learning}
Tanno, R., Saeedi, A., Sankaranarayanan, S., Alexander, D.C., Silberman, N.:
  Learning from noisy labels by regularized estimation of annotator confusion.
  In: Proceedings of the IEEE Conference on Computer Vision and Pattern
  Recognition. pp. 11244--11253 (2019)

\bibitem{wang2012multi}
Wang, H., Suh, J.W., Das, S.R., Pluta, J.B., Craige, C., Yushkevich, P.A.:
  Multi-atlas segmentation with joint label fusion. IEEE transactions on
  pattern analysis and machine intelligence  \textbf{35}(3),  611--623 (2012)

\bibitem{warfield2004simultaneous}
Warfield, S.K., Zou, K.H., Wells, W.M.: Simultaneous truth and performance
  level estimation (staple): an algorithm for the validation of image
  segmentation. IEEE transactions on medical imaging  \textbf{23}(7),  903--921
  (2004)

\bibitem{xu2019whole}
Xu, X., Wang, T., Shi, Y., Yuan, H., Jia, Q., Huang, M., Zhuang, J.: Whole
  heart and great vessel segmentation in congenital heart disease using deep
  neural networks and graph matching. In: International Conference on Medical
  Image Computing and Computer-Assisted Intervention. pp. 477--485. Springer
  (2019)

\bibitem{yaniv2018simpleitk}
Yaniv, Z., Lowekamp, B.C., Johnson, H.J., Beare, R.: Simpleitk image-analysis
  notebooks: a collaborative environment for education and reproducible
  research. Journal of digital imaging  \textbf{31}(3),  290--303 (2018)

\end{thebibliography}




\end{document}